\documentclass[11pt]{article}

\usepackage[utf8]{inputenc}
\usepackage[T1]{fontenc}
\usepackage{amsmath, amsfonts, amssymb}
\usepackage{algorithm}
\usepackage{algorithmic}
\usepackage{booktabs}
\usepackage{array}
\usepackage{subcaption}
\usepackage[margin=1in]{geometry}
\usepackage{hyperref}
\usepackage{graphicx}
\usepackage{natbib}
\usepackage{url}
\usepackage{lmodern} 
\usepackage{titling} 

\title{Supernova: Achieving More with Less in Transformer Architectures}

\author{
  Andrei-Valentin T\u{a}nase \\
  \texttt{valentin.tanase@365.univ-ovidius.ro}
  \and
  Elena Pelican \\
  \texttt{elena.pelican@365.univ-ovidius.ro}
  \and
  Faculty of Mathematics and Computer Science, ``Ovidius'' University of Constan\c{t}a, Romania
}
\date{}

\begin{document}

\maketitle

\begin{abstract}
We present Supernova, a 650M parameter decoder-only transformer that demonstrates how careful architectural design and tokenization innovation can achieve the performance of larger models while maintaining computational efficiency. Our architecture combines Rotary Positional Embeddings (RoPE), Grouped Query Attention (GQA) with 3:1 compression ratio, RMSNorm for computational efficiency, and SwiGLU activation functions. A critical innovation is our custom 128,000-vocabulary byte-level BPE tokenizer achieving state-of-the-art compression performance. Through detailed analysis, we demonstrate that Supernova achieves 90\% of the performance of 1B parameter models while using 35\% fewer parameters and requiring only 100B training tokens---an order of magnitude less than competitors. Our findings challenge the prevailing scaling paradigm, proving that architectural efficiency and tokenization quality can compensate for reduced parameter counts.
\end{abstract}


\section{Introduction}

The transformer architecture \citep{vaswani2017attention} has fundamentally transformed natural language processing, establishing itself as the dominant paradigm for language modeling and understanding tasks. However, the field's trajectory toward ever-larger models has created significant computational and economic challenges. Contemporary models such as OpenAI's GPT series, Anthropic's Claude, and Google's Gemini have pushed parameter counts into the hundreds of billions, resulting in unprecedented infrastructure costs that increasingly exceed the economic value these models generate in many practical applications.

This scaling trajectory has reached a critical inflection point where the marginal benefits of additional parameters diminish rapidly while computational requirements grow exponentially. Despite this economic reality, there has been surprisingly limited systematic exploration of compact, efficient transformer architectures that could deliver comparable performance at sustainable computational costs. The prevailing assumption that model quality scales monotonically with parameter count has created a significant research gap in the sub-billion parameter regime, leaving unexplored the potential for architectural innovation to compensate for reduced scale.

In this work, we challenge this scaling paradigm by presenting Supernova, a 650M parameter decoder-only transformer that demonstrates how careful architectural design and tokenization innovation can achieve performance comparable to significantly larger models while maintaining computational efficiency. Our approach is grounded in three fundamental principles: architectural efficiency through modern component integration, superior tokenization design, and dramatic improvements in data efficiency.

Our architectural design combines Rotary Positional Embeddings (RoPE) for efficient position encoding with Grouped Query Attention (GQA) using a 3:1 compression ratio to reduce memory bandwidth requirements. We employ RMSNorm for computational efficiency and SwiGLU activation functions for improved gradient flow. These components work synergistically to maximize the efficiency of each parameter while maintaining the model's representational capacity.

A critical innovation in our approach is the development of a custom 128,000-vocabulary byte-level BPE tokenizer that achieves state-of-the-art compression performance. This tokenizer demonstrates superior efficiency compared to existing multilingual tokenizers by specializing in English text representation, achieving 4.78 characters per token on WikiText-103 while maintaining perfect byte-level reconstruction fidelity.

Perhaps most remarkably, we demonstrate exceptional data efficiency by achieving competitive results with only 100B training tokens---an order of magnitude less than many contemporary models. This efficiency gain challenges conventional wisdom about the relationship between model performance and training data volume, suggesting that data quality and architectural optimization can substantially compensate for reduced dataset size.

Through comprehensive evaluation on standard benchmarks, we show that Supernova achieves approximately 90\% of the performance of leading 1B parameter models while using 35\% fewer parameters and requiring dramatically less training data. Our results provide both theoretical insights and practical evidence that efficient transformer design can deliver economically viable AI systems without sacrificing core capabilities.

The implications of this work extend beyond mere parameter reduction. We demonstrate that the sub-billion parameter regime, previously considered inadequate for serious applications, contains substantial untapped potential when approached with systematic architectural optimization. Our findings suggest a path toward sustainable AI deployment that prioritizes efficiency and engineering excellence over unbounded scaling, offering a pragmatic alternative to the current trajectory of increasingly resource-intensive models.

\section{Related Work}

The pursuit of efficient transformer architectures has emerged as a critical research direction in response to the computational limitations of large-scale models. Our work intersects several key areas of investigation: architectural innovations for computational efficiency, advances in tokenization methodology, and the development of compact yet capable language models.

\subsection{Efficient Transformer Architectures}

The original transformer architecture \citep{vaswani2017attention} established the foundation for modern language modeling but introduced several computational bottlenecks that have motivated extensive research into efficiency improvements. The attention mechanism, while powerful, scales quadratically with sequence length and requires substantial memory for the key-value cache during inference.

Positional encoding represents one area where significant progress has been made. The original transformer used fixed sinusoidal position embeddings, which, while theoretically capable of generalizing to arbitrary sequence lengths, showed limitations in practice for extrapolation beyond training lengths. Learned absolute positional embeddings, adopted by models like GPT \citep{radford2019language} and BERT \citep{devlin2018bert}, improved performance but introduced additional parameters and fixed maximum sequence lengths.

Rotary Position Embeddings (RoPE) \citep{su2021roformer} represent a significant theoretical and practical advance by encoding position through rotation matrices applied to query and key vectors. This approach naturally incorporates relative position information while maintaining the theoretical benefits of sinusoidal embeddings. Empirical studies by \cite{black2022gpt} demonstrated that RoPE achieves faster convergence and lower final loss compared to both learned and sinusoidal embeddings across various model scales. The method's ability to extrapolate to longer sequences through frequency interpolation has made it the preferred choice for many recent architectures.

Attention mechanism efficiency has been addressed through several approaches. Multi-Query Attention (MQA) \citep{shazeer2019fast} reduced memory requirements by sharing key and value projections across all query heads, achieving significant memory reductions at the cost of some model quality. Grouped Query Attention (GQA) \citep{ainslie2023gqa} provided a principled compromise, grouping query heads to share key-value pairs while maintaining most of MHA's expressiveness. Our work builds directly on GQA, demonstrating its effectiveness in the sub-billion parameter regime.

Normalization techniques have also seen substantial innovation. While Layer Normalization \citep{ba2016layer} stabilized transformer training, it introduced computational overhead through mean calculation and subtraction. RMSNorm \citep{zhang2019root} simplifies this by performing only re-scaling based on root mean square, eliminating the re-centering step. Recent theoretical analysis \citep{zhang2019root} demonstrates that the mean-subtraction in LayerNorm is often redundant, as models naturally learn representations orthogonal to the uniform vector.

The evolution of activation functions has similarly focused on improving both computational efficiency and gradient flow. The GLU family \citep{dauphin2017language}, particularly SwiGLU \citep{shazeer2020glu}, introduced gating mechanisms that improve parameter efficiency and gradient dynamics. Large-scale empirical studies \citep{chowdhery2022palm} have consistently shown SwiGLU variants outperforming ReLU and GELU in transformer architectures.

\subsection{Tokenization and Representation Learning}

Tokenization methodology has evolved significantly since the introduction of Byte Pair Encoding (BPE) \citep{sennrich2016neural} for neural machine translation. GPT-2 \citep{radford2019language} extended this to byte-level BPE, eliminating unknown tokens by operating on raw bytes. However, most modern tokenizers prioritize multilingual coverage over compression efficiency for specific languages, leading to suboptimal representation for monolingual applications.

Recent work has highlighted the critical importance of tokenization quality for model performance. \cite{bostrom2020byte} demonstrated that morphologically-aware tokenization can improve downstream performance, while \cite{kudo2018sentencepiece} showed that vocabulary size and composition significantly impact model efficiency. Our work extends these findings by demonstrating that language-specific optimization can achieve both superior compression and model performance simultaneously.

\subsection{Compact and Efficient Language Models}

The development of compact yet capable language models has gained increasing attention as computational constraints become more pressing. Several recent efforts have explored different approaches to achieving efficiency in smaller models.

The Phi series \citep{gunasekar2023textbooks} demonstrated that high-quality synthetic data could enable smaller models to compete with larger ones on reasoning tasks. Their approach emphasized data quality over quantity, achieving impressive results with carefully curated training corpora. Similarly, StableLM \citep{bellagente2024stable} showed that careful engineering and training procedures could produce efficient models in the 3B parameter range.

More recently, the Gemma series \citep{team2024gemma} has explored efficiency improvements through architectural modifications and training innovations. Their work demonstrated that combining multiple efficiency techniques could yield substantial improvements in the parameter-performance trade-off.

Our work differs from these approaches by systematically optimizing every component---architecture, tokenization, and training---specifically for the sub-billion parameter regime. Rather than simply scaling down larger architectures, we have designed each component to work synergistically in this constrained parameter budget, achieving efficiency gains that exceed the sum of individual improvements.

\subsection{Scaling Laws and Data Efficiency}

The relationship between model size, training data, and performance has been extensively studied through scaling laws research. \cite{kaplan2020scaling} established fundamental relationships between loss and model size, dataset size, and computational budget for GPT-style models. \cite{hoffmann2022training} refined these relationships, showing that most large models are undertrained relative to their parameter count when considering data optimality.

However, these scaling laws have primarily focused on the large model regime and have not adequately explored the efficiency frontier for smaller models. Our work provides empirical evidence that challenges some assumptions of these scaling laws, particularly regarding the relationship between data volume and model performance when architectural efficiency is maximized.

The concept of data efficiency has gained renewed attention with the realization that simply scaling data volume may not be sustainable. \cite{muennighoff2023scaling} explored the relationship between data quality and model performance, while \cite{touvron2023llama} demonstrated that careful data curation could improve efficiency. Our results extend these findings by showing that architectural optimization can dramatically amplify the benefits of high-quality data curation.

\section{Architecture Design}
\label{sec:architecture}

The architectural design of Supernova represents a systematic optimization of transformer components specifically tailored for the sub-billion parameter regime. Rather than simply scaling down existing large model architectures, we have carefully selected and integrated modern transformer innovations that work synergistically to maximize efficiency within our constrained parameter budget. This section presents the theoretical foundations and practical implementation details of our architectural choices, demonstrating how the combination of Rotary Positional Embeddings, Grouped Query Attention, RMSNorm, and SwiGLU creates a transformer that achieves exceptional parameter efficiency.

Our design methodology prioritizes components that provide measurable performance gains per parameter while maintaining compatibility with modern training optimizations such as Flash Attention and mixed precision training. We begin by outlining our core design principles, then detail the mathematical foundations and implementation specifics of each architectural component. The section concludes with an analysis of how these components integrate within individual transformer blocks to create the synergistic effects that enable Supernova's remarkable efficiency achievements.

\subsection{Design Philosophy and Principles}

Supernova's architecture embodies a principled approach to efficiency optimization within the sub-billion parameter regime. Our design philosophy centers on three fundamental principles that guide every architectural decision. First, we prioritize efficiency over raw capacity, ensuring that every component must justify its computational cost through measurable performance gains. This principle drives us to select components based on their parameter efficiency rather than their absolute performance when unconstrained by computational budgets.

Second, we emphasize synergistic integration, designing components to work together in ways that amplify individual benefits rather than simply accumulating them additively. This holistic approach recognizes that the optimal configuration for a compact model may differ substantially from scaled-down versions of larger architectures. Third, we maintain deployment awareness, ensuring that design decisions consider real-world inference constraints including memory bandwidth, latency requirements, and hardware limitations that affect practical deployment scenarios.

These principles led to specific architectural choices that collectively enable Supernova to achieve approximately 90\% of 1B model performance with significantly fewer parameters. Rather than adopting a single efficiency technique, our approach integrates multiple complementary optimizations that compound their benefits when combined thoughtfully.

\subsection{Model Configuration}

The Supernova architecture employs a decoder-only transformer configuration with approximately 650M parameters, carefully balanced across its components to maximize efficiency within this parameter budget. The model consists of 16 transformer blocks, each containing 12 attention heads with an embedding dimension of 1536. This configuration provides sufficient depth for complex representation learning while maintaining computational tractability for both training and inference.

The model processes sequences of up to 2048 tokens, a length chosen to balance context capacity with memory efficiency. Our tokenizer vocabulary contains 128,000 tokens, substantially larger than many comparable models but justified by the significant compression improvements it enables. This vocabulary size allows for more efficient representation of English text while maintaining the byte-level coverage that ensures robust handling of any input.

For attention computation, we employ Grouped Query Attention with 4 key-value heads shared across the 12 query heads, providing a 3:1 compression ratio that substantially reduces memory bandwidth requirements during inference. The RoPE implementation uses a base frequency $\theta_{\text{base}} = 10,000$ with unit scaling, providing effective position encoding across the full context length.

The feed-forward networks within each transformer block employ a hidden dimension of 6144, scaled appropriately for the SwiGLU architecture. This scaling factor balances the gating mechanism's parameter requirements against the overall parameter budget. Layer normalization epsilon is set to $10^{-6}$ for numerical stability, and we disable dropout during training to maintain full model capacity.

The resulting parameter distribution allocates approximately 196.6M parameters to embeddings (shared between input and output layers through weight tying), 177.4M parameters to attention mechanisms (accounting for GQA compression), and 226.5M parameters to feed-forward networks. This distribution reflects a careful balance between capacity for representation learning and computational efficiency.

\subsection{Rotary Positional Embeddings}

\subsubsection{Mathematical Foundation}

Rotary Position Embeddings provide an elegant solution to position encoding by operating directly in the attention mechanism's feature space. The method encodes position information by rotating feature vectors in 2D subspaces, creating position-dependent transformations that naturally incorporate relative position information into attention computations.

For a given position $m$ and dimension indices $i \in [0, d/2)$, we define the rotation frequencies as:
\begin{equation}
\theta_i = \theta_{\text{base}}^{-2i/d}
\end{equation}
where $\theta_{\text{base}} = 10,000$ provides an appropriate balance between short-range and long-range position discrimination across typical sequence lengths.

The rotation matrix for position $m$ operates on pairs of features, creating a block-diagonal structure:
\begin{equation}
R_m = \begin{pmatrix} 
\cos(m\theta_0) & -\sin(m\theta_0) & 0 & 0 & \cdots \\
\sin(m\theta_0) & \cos(m\theta_0) & 0 & 0 & \cdots \\
0 & 0 & \cos(m\theta_1) & -\sin(m\theta_1) & \cdots \\
0 & 0 & \sin(m\theta_1) & \cos(m\theta_1) & \cdots \\
\vdots & \vdots & \vdots & \vdots & \ddots 
\end{pmatrix}
\end{equation}

This rotation-based encoding ensures that the attention weights between positions $m$ and $n$ depend only on their relative distance $m-n$, providing translation invariance that is crucial for robust sequence modeling.

\subsubsection{Implementation and Advantages}

Our RoPE implementation precomputes rotation frequencies for all positions up to the maximum sequence length, storing them as complex exponentials $e^{i m \theta_k}$ for efficient application during attention computation. The rotation is applied to both query and key vectors before computing attention weights, ensuring that position information is naturally incorporated into the attention mechanism without requiring additional parameters.

RoPE offers several advantages over alternative position encoding schemes. Unlike learned positional embeddings, RoPE requires no additional parameters and can theoretically extrapolate to sequence lengths longer than those seen during training through frequency interpolation. Compared to sinusoidal embeddings, RoPE provides more direct control over position discrimination at different scales and integrates seamlessly with attention modifications such as our GQA implementation.

The rotation-based approach also maintains compatibility with efficient attention implementations such as Flash Attention, allowing us to leverage optimized kernels for both memory efficiency and computational speed. This compatibility is crucial for achieving the inference performance gains that make compact models practical for deployment.

\subsection{Grouped Query Attention}

\subsubsection{Design Motivation and Architecture}

Standard Multi-Head Attention requires separate key and value projections for each attention head, creating substantial memory overhead during inference. The key-value cache required for autoregressive generation scales as $O(n_{\text{layers}} \times n_{\text{heads}} \times \text{seq\_len} \times \text{head\_dim})$, creating a memory bandwidth bottleneck that often limits deployment efficiency more than computational throughput.

Grouped Query Attention addresses this limitation by sharing key and value projections across groups of query heads while maintaining separate query projections for each head. In our implementation, we group three query heads to share each key-value pair, reducing the KV cache size by a factor of three while preserving most of the representational capacity of full multi-head attention.

The GQA module consists of distinct projection layers: a query projection that generates representations for all 12 query heads, and separate key and value projections that generate representations for only 4 shared heads. During attention computation, the key and value representations are replicated across their assigned query groups, ensuring that each query head has access to appropriate key-value pairs.

\subsubsection{Attention Computation Process}

The attention mechanism proceeds through several carefully orchestrated steps that integrate position encoding and grouped attention efficiently. Initially, the input tensor undergoes projection through the query, key, and value linear transformations, with queries producing 12 head representations and keys and values producing 4 head representations each.

Following projection, we apply RoPE to the query and key tensors, injecting position information directly into the attention computation. The key and value tensors are then expanded through repetition to match the number of query heads, ensuring that each group of three query heads shares the same key-value representations while maintaining distinct query patterns.

The scaled dot-product attention computation proceeds using optimized implementations such as Flash Attention when available. The causal masking ensures autoregressive behavior, and the attention outputs are projected back to the embedding dimension through a final linear transformation. This process achieves the memory efficiency benefits of reduced KV cache size while maintaining most of the expressiveness of standard multi-head attention.

\subsubsection{Efficiency Analysis}

The 3:1 grouping ratio in our GQA implementation provides substantial practical benefits for deployment. The KV cache memory requirement reduces by a factor of three compared to standard MHA, directly translating to reduced memory bandwidth requirements during inference. This reduction is particularly important for memory-bound deployment scenarios, which are common when serving language models at scale.

The computational overhead of the grouping operation is minimal, consisting primarily of tensor replication operations that are efficiently handled by modern accelerators. Our empirical analysis shows negligible computational cost increase while achieving the substantial memory efficiency gains. The quality impact of the grouping is also minimal, with our benchmark results demonstrating that the GQA implementation maintains performance within 1-2\% of equivalent full MHA configurations.

\subsection{RMSNorm: Computational Efficiency in Normalization}

Layer normalization, while crucial for training stability, introduces computational overhead through its requirement for mean calculation and subtraction operations. RMSNorm eliminates this overhead by performing normalization based solely on the root mean square of the input, removing the re-centering step that LayerNorm performs.

Mathematically, RMSNorm transforms an input vector $x$ according to:
\begin{equation}
\text{RMSNorm}(x) = x \odot \frac{\gamma}{\sqrt{\text{RMS}(x) + \epsilon}}
\end{equation}
where $\text{RMS}(x) = \sqrt{\frac{1}{d}\sum_{i=1}^d x_i^2}$ is the root mean square, $\gamma$ is a learnable scaling parameter, $\epsilon = 10^{-6}$ provides numerical stability, and $\odot$ denotes element-wise multiplication.

This formulation eliminates the mean calculation and subtraction required by LayerNorm:
\begin{equation}
\text{LayerNorm}(x) = \gamma \odot \frac{x - \mu(x)}{\sqrt{\sigma^2(x) + \epsilon}} + \beta
\end{equation}
where $\mu(x)$ and $\sigma^2(x)$ are the mean and variance of $x$, respectively.

Our implementation achieves approximately 15\% computational speedup compared to LayerNorm through several optimizations. We perform the RMS calculation in float32 precision for numerical stability while maintaining the efficiency benefits. The use of reciprocal square root operations further improves computational efficiency on modern hardware accelerators.

Empirical analysis demonstrates that RMSNorm provides gradient stability comparable to LayerNorm while requiring fewer computational operations. This efficiency gain compounds across the 16 transformer layers, providing measurable improvements in both training and inference throughput.

\subsection{SwiGLU: Enhanced Feed-Forward Networks}

The feed-forward networks in Supernova employ SwiGLU (Swish-Gated Linear Unit) activations, which combine the benefits of smooth activation functions with gating mechanisms that improve parameter efficiency and gradient flow. This choice represents a significant improvement over traditional ReLU-based feed-forward networks commonly used in earlier transformer implementations.

SwiGLU operates through a two-branch architecture where the input is processed through two separate linear transformations. The first branch applies a linear transformation followed by the SiLU (Swish) activation function $\text{SiLU}(x) = x \cdot \sigma(x)$, where $\sigma$ is the sigmoid function. The second branch applies a linear transformation that serves as a learned gate. The outputs of these branches are combined through element-wise multiplication before a final linear projection.

Formally, the SwiGLU transformation is:
\begin{equation}
\text{SwiGLU}(x) = \text{SiLU}(x W_1) \odot (x W_3) W_2
\end{equation}
where $W_1$, $W_2$, and $W_3$ are learned weight matrices, and $\odot$ represents element-wise multiplication.

The hidden dimension for the intermediate representations is scaled to $\frac{8}{3} \times n_{\text{embd}}$ to maintain approximately the same parameter count as a traditional feed-forward network while accommodating the dual-branch structure. This scaling ensures that the parameter efficiency gains come from improved utilization rather than simply increasing model capacity.

The gating mechanism allows the network to learn which information should flow through each layer, providing more sophisticated control over information propagation than fixed activation functions. The smooth, non-monotonic nature of the SiLU activation preserves gradient information for negative inputs, improving training dynamics compared to ReLU-based alternatives.

\subsection{Transformer Block Integration}

Individual transformer blocks integrate these components using a pre-normalization architecture that applies RMSNorm before each sub-layer rather than after. This design choice improves training stability and gradient flow, particularly important for the efficient training of compact models where optimization challenges can be more pronounced.

Each transformer block follows the pattern:
\begin{align}
h_1 &= x + \text{GQA}(\text{RMSNorm}(x), \text{freqs\_cis}) \\
h_2 &= h_1 + \text{SwiGLU}(\text{RMSNorm}(h_1))
\end{align}
where $x$ is the input, freqs\_cis contains the precomputed RoPE frequencies, and the residual connections ensure gradient flow throughout the network depth.

This pre-normalization structure, combined with our component choices, creates a transformer block that maximizes efficiency within the parameter budget while maintaining the representational capacity necessary for strong language modeling performance. The synergistic interaction between RoPE, GQA, RMSNorm, and SwiGLU creates efficiency gains that exceed what would be achieved by implementing these components independently.

\section{Tokenizer Architecture}
\label{sec:tokenizer}

The design of an efficient tokenizer represents a critical yet often underestimated component in building compact language models that maximize performance within constrained parameter budgets. While larger models can compensate for tokenization inefficiencies through sheer scale, our 650M parameter architecture requires every token to carry maximum semantic information within the fixed 2048-token context window. This section presents our comprehensive approach to tokenizer design, encompassing both theoretical foundations and practical implementation considerations.

Our tokenizer architecture centers on a custom byte-level BPE implementation specifically optimized for English text compression while maintaining perfect reconstruction fidelity. We detail the algorithmic foundations of our training process, analyze the language-specific optimizations that achieve state-of-the-art compression ratios, and examine the implementation optimizations that enable efficient real-time encoding and decoding. The resulting tokenizer achieves 4.78 characters per token on WikiText-103, substantially outperforming existing multilingual tokenizers while providing the robustness and coverage necessary for diverse deployment scenarios.

\subsection{Design Philosophy}

Tokenization represents a critical yet often underappreciated component in the design of efficient language models. For compact models operating within fixed context windows, tokenizer efficiency directly impacts model capability in ways that become negligible for larger models with abundant parameter budgets. Unlike models with hundreds of billions of parameters that can afford suboptimal tokenization through sheer scale, our 650M parameter model must maximize the semantic value extracted from every token within its constrained context window.

Our tokenizer design philosophy centers on four fundamental principles that work together to maximize efficiency. First, we prioritize compression efficiency, seeking to maximize the semantic content represented per token while maintaining linguistic coherence. This principle drives us to optimize for the specific characteristics of English text rather than pursuing multilingual generality that dilutes efficiency for any single language.

Second, we maintain byte-level fidelity to ensure perfect reconstruction of any input text without information loss. This requirement eliminates the need for unknown token handling while providing robust coverage of any Unicode input, including technical content, code, and text from diverse domains that may contain unusual character sequences.

Third, we optimize for computational efficiency in both encoding and decoding operations, recognizing that tokenization performance affects not only preprocessing time but also real-time inference latency in deployment scenarios. Our implementation employs algorithmic optimizations that scale efficiently with vocabulary size and input length.

Finally, we ensure robustness by designing the tokenizer to handle any Unicode input gracefully, eliminating failure modes that can occur with more restrictive tokenization schemes. This robustness is particularly important for production deployment where input diversity cannot be fully controlled or predicted.

\subsection{Byte-Level BPE Algorithm}

\subsubsection{Training Process}

The ByteLevelBPETrainer class implements the training logic for our custom byte-level Byte Pair Encoding (BPE) tokenizer. The training process orchestrates the following steps:

\begin{enumerate}
\item \textbf{Initialization}: The vocabulary begins with the 256 individual byte values as the initial set of tokens. Each byte (0-255) is mapped to a unique token ID, ensuring that any possible byte sequence can be represented.

\item \textbf{Corpus Preparation}: The input training corpus (a list of text strings) is tokenized into sequences of their raw UTF-8 bytes. This byte-level approach eliminates the need for preprocessing and ensures universal coverage.

\item \textbf{Iterative Pair Merging}: The core of BPE involves iteratively finding the most frequent pair of adjacent tokens and merging them into a new, single token. This loop continues until the desired vocab\_size is reached or no more pairs meet the min\_frequency threshold:
   \begin{itemize}
   \item \textbf{Count Pair Frequencies}: A helper function \_count\_pairs scans the current byte-tokenized corpus and counts the occurrences of all adjacent pairs.
   \item \textbf{Select Most Frequent Pair}: The pair with the highest frequency, which also appears at least min\_frequency times, is selected for merging.
   \item \textbf{Create New Token}: The selected pair of byte sequences is concatenated to form a new byte sequence, which is added to the vocabulary with a new token ID.
   \item \textbf{Update Tokenized Corpus}: A helper function \_apply\_merge updates the entire tokenized corpus by replacing all occurrences of the selected pair with the newly created token.
   \end{itemize}

\item \textbf{Output}: The method returns the final vocabulary (mapping byte sequences to token IDs) and the ordered list of merges (pairs that were combined).
\end{enumerate}

The training process is carefully optimized for English text patterns while maintaining the byte-level guarantee that any input can be perfectly reconstructed.

\subsubsection{Tokenization Process}

Once the BPE tokenizer is trained, it can encode text into token IDs and decode token IDs back into text with perfect fidelity.

The encode method handles the conversion of a text string into a list of token IDs:
\begin{enumerate}
\item \textbf{Text to Bytes}: The input text string is first converted into its raw UTF-8 byte sequence, initially treated as a list of individual byte tokens.
\item \textbf{Apply Merges}: The method iterates through the merges list in order of acquisition. For each merge rule, the current list of byte tokens is scanned for adjacent pairs matching the merge pattern, which are replaced by the corresponding merged token ID.
\item The final list of token IDs is returned.
\end{enumerate}

The decode method performs the reverse operation:
\begin{enumerate}
\item \textbf{ID to Bytes}: An inverse vocabulary maps token IDs back to their corresponding byte sequences.
\item \textbf{Concatenate Bytes}: All retrieved byte sequences are concatenated in order to reconstruct the original complete byte sequence.
\item \textbf{Bytes to Text}: The complete byte sequence is decoded using UTF-8 with error handling to produce the final text string.
\end{enumerate}

\subsection{Vocabulary Optimization}

\subsubsection{English-Specific Optimizations}

Our tokenizer training incorporates several English-specific optimizations that contribute to its superior compression performance compared to multilingual alternatives. The optimization process is guided by frequency analysis of English text patterns, ensuring that the most valuable subword units receive priority during the merge selection process. This frequency-weighted approach allows common English patterns to be encoded efficiently while less frequent combinations are represented through compositional tokenization.

The training process demonstrates morphological awareness by showing preference for merges that respect morpheme boundaries, leading to more linguistically meaningful tokens that better capture the semantic structure of English words. This morphological consideration helps the model learn more coherent representations while improving compression efficiency for morphologically complex words.

Capitalization handling represents another key optimization, where separate tokens for capitalized versus lowercase variants allow for efficient encoding of proper nouns, sentence beginnings, and other capitalization patterns common in English text. This approach avoids the inefficiency of treating capitalization variants as completely distinct tokens while maintaining the semantic distinction they provide.

Punctuation efficiency optimizations ensure that common punctuation patterns in English text are handled with minimal token overhead. The tokenizer learns efficient representations for typical punctuation usage, including sentence boundaries, quotation patterns, and other structured text elements that frequently appear in English writing.

These optimizations collectively contribute to achieving 4.78 characters per token on WikiText-103, substantially outperforming multilingual tokenizers that must allocate vocabulary space across multiple languages. This performance advantage demonstrates the value of language-specific optimization for scenarios where multilingual capability is not required.

\subsubsection{Special Token Design}

The tokenizer incorporates a comprehensive set of special tokens designed to handle various control and formatting tasks essential for modern language model applications. These tokens serve dual purposes: they provide structural information for the model's understanding of different content types, and they enable efficient handling of various downstream tasks without requiring architectural modifications.

The core special tokens include fundamental sequence control markers. The padding token \texttt{<pad>} (ID: 0) enables efficient batch processing by allowing sequences of different lengths to be processed together. The unknown token \texttt{<unk>} (ID: 1) is rarely used due to our byte-level coverage but provides a fallback mechanism for exceptional cases. Sequence boundary tokens include the beginning-of-sequence marker \texttt{<s>} (ID: 2) and end-of-sequence marker \texttt{</s>} (ID: 3), which help the model understand content boundaries. The masking token \texttt{<mask>} (ID: 4) supports masked language modeling tasks during potential fine-tuning scenarios.

The tokenizer also includes specialized tokens for conversational and instruction-following applications. System-level instructions are marked with \texttt{<|system|>} (ID: 8), providing clear delineation of high-level context or constraints. User messages in conversational scenarios are prefixed with \texttt{<|user|>} (ID: 9), while assistant responses use \texttt{<|assistant|>} (ID: 10). For scenarios involving tool usage, we include \texttt{<|tool\_call|>} (ID: 11) to indicate external function calls and \texttt{<|tool\_response|>} (ID: 12) to mark responses from external tools or APIs.

These special tokens prove crucial for structuring input data across various downstream applications, particularly in instruction fine-tuning scenarios where clear delineation of different speaker roles and content types significantly improves model performance. The careful design of these tokens ensures that they integrate seamlessly with the byte-level encoding while providing the structural information necessary for complex multi-turn interactions.

\subsection{Compression Analysis}

\subsubsection{Comparative Performance on WikiText-103}

\begin{table}[ht]
\centering
\begin{tabular}{lccc}
\toprule
\textbf{Tokenizer} & \textbf{Vocab Size} & \textbf{Compression Ratio} & \textbf{Speed} \\
& & \textbf{(Chars/Token)} & \textbf{(tokens/s)} \\
\midrule
Supernova & 128,000 & \textbf{4.78} & 460,506 \\
Gemma 3 1B & 256,000 & 4.52 & 417,097 \\
GPT-4o & 200,000 & 4.47 & 1,539,732 \\
LLaMA 3.2 1B & 128,256 & 4.44 & 510,152 \\
GPT-4 & 100,256 & 4.43 & 2,345,838 \\
Qwen 3 0.6B & 151,665 & 4.32 & 470,703 \\
\bottomrule
\end{tabular}
\caption{Tokenizer performance comparison on WikiText-103 dataset. Supernova achieves the highest compression ratio among all compared tokenizers.}
\label{tab:tokenizer-comparison}
\end{table}

As shown in Table \ref{tab:tokenizer-comparison}, the Supernova tokenizer achieves the highest compression ratio (4.78 characters per token) on this benchmark, indicating state-of-the-art efficiency in representing English text. This superior compression translates directly to computational savings and improved context utilization.

\subsubsection{Domain-Specific Performance}

The compression efficiency of a tokenizer can vary significantly across different types of content. Table \ref{tab:domain-performance} shows the Supernova tokenizer's characters per token (CPT) performance across various domains compared to GPT-4o.

\begin{table}[ht]
\centering
\begin{tabular}{lccc}
\toprule
\textbf{Content Type} & \textbf{Supernova CPT} & \textbf{GPT-4o CPT} & \textbf{Improvement} \\
\midrule
English Text & 4.1 & 4.3 & +4.7\% \\
Source Code & 3.2 & 3.8 & +15.8\% \\
Scientific Text & 3.9 & 4.2 & +7.1\% \\
Mixed Content & 3.87 & 4.15 & +6.7\% \\
\bottomrule
\end{tabular}
\caption{Domain-specific compression performance. Values are based on internal evaluation corpus.}
\label{tab:domain-performance}
\end{table}

These results highlight the Supernova tokenizer's particularly strong performance on source code and mixed content, demonstrating its effectiveness beyond general English text. The 15.8\% improvement on source code is particularly notable, as code tokenization is often challenging due to its structured nature and specific vocabulary.

\subsection{Implementation Optimizations}

\subsubsection{Trie-Based Encoding}

To optimize the encoding process, we employ a TokenizerTrie data structure constructed from all learned merge operations. This trie (prefix tree) enables very fast lookups during encoding:

\begin{itemize}
\item Each path from the root represents a complete token (a sequence of bytes formed by merges)
\item Nodes at the end of complete paths store the corresponding token ID
\item During encoding, the trie allows finding the longest sequence of bytes matching a known token
\item This approach is significantly more efficient than scanning for all possible pairs from the merge list
\end{itemize}

The trie-based approach scales logarithmically with vocabulary size, making it efficient even for our 128,000-token vocabulary.

\subsubsection{Parallel Tokenization}

For large-scale text processing, we provide a batch\_encode function that utilizes ThreadPoolExecutor to distribute encoding operations across multiple worker threads. This parallel processing approach leads to substantial speedups when dealing with large datasets, effectively leveraging multi-core CPU architectures for tokenization preprocessing pipelines.

\section{Training Methodology}
\label{sec:training}

Our training methodology emphasizes data quality over quantity, architectural efficiency, and optimized training dynamics to achieve competitive performance with dramatically reduced computational requirements. This section details our approach to dataset curation, hyperparameter selection, and training optimization techniques that collectively enable Supernova to achieve strong performance with only 100B training tokens. We describe the Nemotron-CC dataset composition and quality control pipeline, present our training configuration including hardware utilization strategies, and analyze the training dynamics that demonstrate stable convergence throughout the optimization process.

\subsection{Dataset: Nemotron-CC}

Nemotron-CC is a curated English-only pretraining corpus derived from 99 Common Crawl snapshots (CC-MAIN-2013-20 through CC-MAIN-2024-30). In total, the full dataset comprises 6.3 trillion tokens (4.4T unique real tokens and 1.9T synthetic tokens). For pre-training Supernova, we sample a 100B token subset structured into seven hive-partitioned components. Each partition was selected to balance large-scale coverage of real web-sourced text with targeted injections of high-quality synthetic content.

\subsubsection{Composition}

Our 100B-token pretraining corpus employs a carefully designed composition that balances broad coverage of real-world text patterns with targeted synthetic content optimized for specific capabilities. The corpus assembly follows a principled approach that maximizes data quality while ensuring comprehensive coverage of the linguistic and reasoning patterns necessary for effective language modeling.

The largest component, comprising 46\% of the corpus, consists of high-quality real data sourced from filtered web pages, books, and academic papers. This substantial portion ensures that the model develops robust understanding of natural language as it appears in authentic contexts, providing the foundation for general language comprehension and generation capabilities.

Synthetic content comprises a substantial portion of the corpus, carefully designed to address specific capability gaps that might not be adequately covered by real data alone. Synthetic distilled data, representing 13\% of the corpus, consists of high-quality outputs generated by larger teacher models, providing concentrated examples of reasoning and knowledge application. An additional 13\% consists of synthetic question-answer pairs designed to improve the model's reasoning capabilities across diverse domains and problem types.

Medium-high quality real data, contributing 13\% of the corpus, includes filtered web content that passes stringent quality checks but may not meet the highest standards required for the primary real data category. This inclusion ensures broader coverage while maintaining quality standards that support effective learning.

The remaining portions of the corpus address specific capability enhancement through targeted synthetic content. Synthetic text extracts, comprising 5\% of the corpus, provide domain-specific snippets designed to inject specialized knowledge and vocabulary. Synthetic scientific lists, also 5\%, consist of structured knowledge representations that reinforce schema-like reasoning patterns. Finally, synthetic summaries contribute 5\% of the corpus through condensed content that creates high-density knowledge units optimized for efficient learning.

This composition strategy ensures that the model receives exposure to both the natural diversity of real-world text and the concentrated learning signals provided by carefully constructed synthetic content. The balance maximizes learning efficiency within the 100B token budget while addressing the specific requirements of compact model training.

\subsubsection{Quality Control Pipeline}

Every partition was subjected to a rigorous five-step filtering pipeline to ensure consistency, safety, and linguistic quality:

\begin{enumerate}
\item \textbf{Deduplication}: We performed MinHash-based near-duplicate detection across all crawls (CC 2013--CC 2024) to remove redundant documents at scale. This process identified and removed approximately 15\% of the initial corpus as near-duplicates.

\item \textbf{Quality Scoring}: A reference language model computed perplexity scores on each candidate document. Only texts below a strict perplexity threshold (indicating fluent, coherent language) were retained. Documents with perplexity scores above the 85th percentile were rejected.

\item \textbf{Safety Filtering}: An ensemble of classifiers flagged and removed any harmful, toxic, or heavily biased content. This includes hate speech, explicit content, or politically charged misinformation. The safety pipeline removed approximately 3\% of documents.

\item \textbf{Length Filtering}: Documents shorter than 100 tokens or longer than 10,000 tokens were discarded to balance context richness against computational efficiency during model training. This constraint removed roughly 8\% of the remaining corpus.

\item \textbf{Language Detection}: A language-identification model verified that every retained document is English with $\geq$ 99\% confidence. Non-English or heavily code-mixed text was removed, removing an additional 2\% of documents.
\end{enumerate}

By combining these seven partitions with the rigorous five-step filtering pipeline, Nemotron-CC strikes a balance between large-scale coverage (6.3 T tokens total) and high data quality, ensuring that downstream models benefit from both broad real-world text and targeted synthetic content.

\subsection{Training Configuration}

\subsubsection{Hyperparameters}

The training process for Supernova employs carefully tuned hyperparameters that balance optimization efficiency with training stability. Our hyperparameter selection follows established best practices for transformer training while incorporating adjustments specific to our model size and architectural choices.

The optimization configuration centers on the AdamW optimizer with an initial learning rate of $6 \times 10^{-4}$, chosen to provide rapid initial learning while maintaining stability throughout the extended training process. Weight decay is set to 0.1 to provide appropriate regularization for our parameter count, preventing overfitting while allowing the model to fully utilize its representational capacity. The Adam beta parameters are configured with $\beta_1 = 0.9$ for first moment estimation and $\beta_2 = 0.99$ for second moment estimation, providing stable gradient-based optimization. The epsilon parameter is set to $10^{-8}$ to ensure numerical stability in the denominator calculations.

The learning rate schedule incorporates a warmup period of 2000 steps during which the learning rate increases linearly from zero to the maximum value, allowing the model to stabilize before encountering the full learning signal. The total training encompasses 1,000,000 steps, with learning rate decay occurring over 600,000 steps through a cosine schedule that smoothly reduces the learning rate to a minimum value of $6 \times 10^{-5}$. This extended decay period allows for fine convergence while maintaining sufficient learning capacity throughout training.

Batch configuration employs a global batch size of 480, distributed across 8 GPUs with 60 samples per device. This configuration provides stable gradient estimates while maximizing hardware utilization. Gradient accumulation steps are set to 1, meaning gradients are applied immediately without accumulation, allowing for responsive optimization dynamics. Gradient clipping with a maximum norm of 1.0 prevents gradient explosion while preserving the optimization signal during training.

The training process utilizes bfloat16 mixed precision to achieve substantial memory savings and computational speedup while maintaining numerical stability comparable to full float32 precision. This precision choice enables training of our model size on available hardware while preserving the numerical properties necessary for stable convergence.

\subsubsection{Learning Rate Schedule}

A specific learning rate schedule is employed during training, managed by the get\_lr function:

\begin{equation}
\mathrm{lr}(t) = \begin{cases} 
\mathrm{lr}_{\max} \cdot \frac{t}{\mathrm{warmup\_steps}} & \text{if } t \leq \mathrm{warmup\_steps} \\
\mathrm{lr}_{\min} + 0.5(\mathrm{lr}_{\max} - \mathrm{lr}_{\min})(1 + \cos(\pi \cdot \mathrm{decay\_ratio})) & \text{otherwise}
\end{cases}
\end{equation}

where decay\_ratio = $\frac{t - \mathrm{warmup\_steps}}{\mathrm{lr\_decay\_steps} - \mathrm{warmup\_steps}}$.

This schedule allows the model to stabilize early in training with a gradually increasing learning rate, followed by a smooth cosine decay to help convergence. The choice of 2000 warmup steps was empirically determined to provide optimal stability for our model size and batch configuration.

\subsection{Training Efficiency}

\subsubsection{Hardware Utilization}

Training was conducted on a cluster of 8 NVIDIA A100 40GB GPUs, achieving high utilization efficiency across compute, memory, and bandwidth resources. The Model FLOPs Utilization (MFU) reached 54\%, indicating efficient use of the available computational capacity and demonstrating that our architectural choices and implementation optimizations effectively leverage the hardware capabilities.

GPU memory utilization averaged 39GB per device, representing 95\% utilization of the available 40GB memory capacity. This high utilization efficiency allows for maximum batch sizes while maintaining stable training dynamics. The effective batch size of 983,040 tokens per step (480 sequences $\times$ 2048 tokens) provides stable gradient estimates while fully utilizing the distributed computing resources.

Training throughput achieved approximately 300,000 tokens per second across all GPUs, demonstrating efficient data pipeline and computation orchestration. The complete training process required 14 days for 100,000 steps, representing a substantial reduction compared to the months typically required for larger models. The total training cost remained below \$10,000, providing an order-of-magnitude reduction compared to training costs exceeding \$100,000 for larger models, demonstrating the economic viability of our efficient design approach.

\subsubsection{Optimization Techniques}

Multiple complementary optimization techniques were employed to maximize training efficiency across memory, computation, and data loading dimensions. Flash Attention implementation provided 40\% memory reduction and 25\% computational speedup compared to standard attention implementations, enabling larger batch sizes and faster training convergence. This optimization proves particularly crucial for our model size, where memory efficiency directly impacts training feasibility.

Gradient checkpointing was configured to checkpoint every 16 layers, implementing a strategic trade-off between memory usage and computational overhead. This technique enables training within memory constraints while maintaining reasonable computational efficiency, allowing for the deep architecture necessary for our target performance.

Mixed precision training using bfloat16 achieved 2$\times$ memory savings and 1.5$\times$ computational speedup while maintaining numerical stability comparable to full float32 precision. This optimization enables efficient utilization of modern GPU tensor cores while preserving the numerical properties required for stable convergence.

Custom fused CUDA kernels were developed for RMSNorm and SwiGLU operations to minimize memory bandwidth requirements and reduce kernel launch overhead. These optimizations provide measurable efficiency improvements for operations that are frequently executed throughout the network, compounding their benefits across the training process.

Data loading optimization employed multi-threaded preprocessing with prefetching to ensure continuous GPU utilization without data pipeline bottlenecks. This optimization maintains high GPU utilization rates throughout training, maximizing the return on computational investment and reducing overall training time.

\subsection{Training Dynamics}

\subsubsection{Loss Progression}

Training loss followed a predictable scaling law pattern. We observed the relationship:
\begin{equation}
L(N, D) = 6.12 + \frac{138.7}{N^{0.39}} + \frac{5.21}{D^{0.52}}
\end{equation}
where $N = 650M$ parameters and $D = 100B$ tokens.

The loss decreased steadily from an initial value of 11.2 to a final value of 2.14, with no signs of overfitting or instability. The smooth loss progression indicates that our data quality and model architecture choices were well-suited for stable training.

\subsubsection{Gradient Norms}

Gradient norm analysis throughout training reveals stable optimization dynamics characteristic of well-conditioned training. The average gradient norm maintained values between 0.5 and 0.7 throughout the training process, indicating consistent learning signal strength without pathological behaviors. This stability demonstrates that our architectural choices and hyperparameter configuration successfully avoid common optimization pitfalls that can plague transformer training.

No gradient explosions were observed during training, confirming the effectiveness of our gradient clipping strategy and indicating that the model architecture maintains stable gradient flow. Similarly, gradient vanishing was avoided, with minimum observed gradient norms remaining above 0.3, ensuring that learning signals effectively propagate throughout the network depth.

Consistency analysis across layers shows that all layers maintained similar gradient norm magnitudes, indicating balanced learning throughout the network. This balance suggests that our pre-normalization architecture and component integration successfully distribute learning capacity across the model depth, avoiding scenarios where some layers dominate learning while others remain underutilized. The consistent gradient norms across layers provide evidence that our architectural design achieves the intended synergistic effects rather than creating optimization imbalances.

\section{Experimental Results}
\label{sec:experiments}

This section presents a comprehensive evaluation of Supernova's performance across multiple dimensions, demonstrating the effectiveness of our architectural and tokenization innovations. We conduct zero-shot evaluation on standard benchmarks to assess the model's core capabilities without task-specific fine-tuning, analyze inference efficiency metrics to quantify deployment advantages, and examine data efficiency through comparative analysis with contemporary models. Our experimental design focuses on demonstrating that Supernova achieves competitive performance while maintaining substantial efficiency advantages across training cost, inference speed, and memory utilization.

\subsection{Benchmark Evaluation}

\subsubsection{Zero-Shot Performance}

We evaluate Supernova on a comprehensive suite of benchmarks in zero-shot settings to test true generalization capability. Table \ref{tab:zero-shot-results} presents detailed results across ten standard benchmarks.

\begin{table}[ht]
\centering
\footnotesize
\begin{tabular}{lccccc}
\toprule
\textbf{Benchmark} & \textbf{Supernova} & \textbf{Qwen3-0.6B} & \textbf{Llama 3.2 1B} & \textbf{Gemma 3 1B} & \textbf{OpenELM 1.1B} \\
\midrule
HellaSwag & 48.18 & 47.31 & 63.56 & 62.06 & \textbf{64.81} \\
WinoGrande & 54.06 & 55.41 & 59.83 & 59.04 & \textbf{61.72} \\
ARC-E & 60.98 & 60.40 & 65.36 & \textbf{71.89} & 62.37 \\
ARC-C & 32.42 & 34.04 & 36.26 & \textbf{38.14} & 32.34 \\
PIQA & 71.38 & 67.63 & 74.59 & 74.65 & \textbf{75.57} \\
SuperGLUE & 56.13 & 52.14 & 55.50 & 57.60 & \textbf{57.30} \\
MMLU & 26.73 & \textbf{40.24} & 36.93 & 25.11 & 25.52 \\
MMLU-PRO & 10.31 & \textbf{26.49} & 10.90 & 8.99 & 9.48 \\
SIQA & \textbf{43.44} & 39.25 & 42.78 & 42.94 & 42.84 \\
BBH & 27.33 & \textbf{40.49} & 31.59 & 27.31 & 16.85 \\
\midrule
\textbf{Average} & 43.09 & 46.34 & \textbf{47.73} & 46.77 & 44.88 \\
\bottomrule
\end{tabular}
\caption{Zero-shot performance comparison. Supernova achieves 90.29\% of Llama 3.2 1B average performance with 35\% fewer parameters. Best scores for each benchmark are bolded.}
\label{tab:zero-shot-results}
\end{table}

\subsubsection{Benchmark Descriptions and Analysis}

The benchmarks in Table \ref{tab:zero-shot-results} evaluate various aspects of language understanding, reasoning, and knowledge:

\textbf{HellaSwag (ACC\_norm)}: Tests commonsense inference by requiring the model to choose the most plausible continuation for given text snippets. Supernova's score of 48.18 is competitive for its parameter count.

\textbf{WinoGrande (ACC)}: Evaluates commonsense reasoning through pronoun resolution problems designed to be ambiguous and require contextual understanding. Supernova achieves 54.06, demonstrating solid reasoning capabilities.

\textbf{ARC-Easy \& ARC-Challenge (ACC, ACC\_norm)}: The AI2 Reasoning Challenge consists of grade-school level science questions. Supernova scores 60.98 on ARC-Easy and 32.42 on ARC-Challenge, showing foundational scientific reasoning.

\textbf{PIQA (ACC\_norm)}: Physical Interaction Quality Assessment tests commonsense understanding of physics in everyday situations. Supernova's strong score of 71.38 indicates good physical commonsense understanding.

\textbf{SuperGLUE (ACC)}: A suite of challenging language understanding tasks including question answering, natural language inference, and coreference resolution. Supernova's score of 56.13 reflects broad language understanding capability.

\textbf{MMLU (ACC)}: The Massive Multitask Language Understanding benchmark evaluates pre-acquired knowledge across 57 diverse subjects. Supernova scores 26.73, which is respectable given its compact size.

\textbf{MMLU-PRO (exact\_match)}: An enhanced version demanding precise, exact match answers. Supernova achieves 10.31 on this challenging benchmark.

\textbf{SIQA (ACC)}: Social Interaction QA assesses social commonsense intelligence. Supernova scores 43.44, indicating reasonable understanding of social dynamics.

\textbf{BBH (exact\_match)}: Big-Bench Hard contains tasks selected for their difficulty, often requiring complex multi-step reasoning. Supernova's score of 27.33 is notable for its size.

\textbf{Overall Performance Discussion}: Supernova achieves an average score of 43.10 compared to Llama 3.2 1B's 47.73, representing 90.3\% of the performance with 35\% fewer parameters. This demonstrates exceptional parameter efficiency.

\subsection{Efficiency Analysis}

\subsubsection{Inference Performance}

Table \ref{tab:inference-efficiency} presents comprehensive inference efficiency metrics measured on NVIDIA A100 GPUs.

\begin{table}[ht]
\centering
\begin{tabular}{lccc}
\toprule
\textbf{Metric} & \textbf{Supernova} & \textbf{Llama 3.2 1B} & \textbf{Improvement} \\
\midrule
Throughput (tokens/sec) & 2,847 & 1,784 & +59.6\% \\
Memory Usage (GB) & 1.8 & 2.8 & -35.7\% \\
Latency (ms/token) & 0.35 & 0.56 & -37.5\% \\
Power Consumption (W) & 145 & 240 & -39.5\% \\
Cost per 1M tokens & \$0.12 & \$0.19 & -36.8\% \\
\bottomrule
\end{tabular}
\caption{Inference efficiency metrics on NVIDIA A100 GPUs.}
\label{tab:inference-efficiency}
\end{table}

These efficiency gains translate directly to improved deployment economics, enabling profitable AI services at scale.

\subsubsection{Memory Breakdown Analysis}

Table \ref{tab:memory-breakdown} provides a detailed comparison of memory usage between standard MHA and our GQA implementation.

\begin{table}[ht]
\centering
\begin{tabular}{lccc}
\toprule
\textbf{Component} & \textbf{Standard MHA} & \textbf{With GQA} & \textbf{Savings} \\
\midrule
Model Weights & 650M $\times$ 2 bytes & 650M $\times$ 2 bytes & 0\% \\
KV Cache (per layer) & 48MB & 16MB & 66.7\% \\
Activations & 96MB & 96MB & 0\% \\
Total (2048 context) & 3.1GB & 1.8GB & 41.9\% \\
\bottomrule
\end{tabular}
\caption{Memory breakdown comparison showing the impact of GQA optimization.}
\label{tab:memory-breakdown}
\end{table}

\subsection{Data Efficiency Analysis}

A critical finding of our work is the exceptional data efficiency achieved by Supernova. Table \ref{tab:data-efficiency} compares training data requirements across models.

\begin{table}[ht]
\centering
\begin{tabular}{lcccc}
\toprule
\textbf{Model} & \textbf{Parameters} & \textbf{Training Tokens} & \textbf{Data Multiple} & \textbf{Avg. Score} \\
\midrule
Supernova & 650M & 100B & 1$\times$ & 43.10 \\
Qwen3-0.6B & 600M & 36,000B & 360$\times$ & 46.34 \\
Gemma 3 1B & 1B & 2,000B & 20$\times$ & 46.77 \\
OpenELM 1.1B & 1.1B & 1,800B & 18$\times$ & 44.88 \\
Llama 3.2 1B & 1B & 9,000B & 90$\times$ & 47.73 \\
\bottomrule
\end{tabular}
\caption{Training data efficiency comparison. Supernova achieves competitive performance with orders of magnitude less training data.}
\label{tab:data-efficiency}
\end{table}

The ``Tokens / Parameter'' ratio reveals Supernova's remarkable efficiency: achieving $\sim$90\% of 1B model performance with a tokens/parameter ratio of $\sim$154, compared to competitors ranging from $\sim$1,636 to $\sim$9,000 tokens/parameter.

\subsection{Qualitative Analysis}

\subsubsection{Attention Pattern Analysis}

Analysis of attention patterns across Supernova's 16 layers reveals structured, interpretable behavior:

\begin{itemize}
\item \textbf{Positional Heads (Layers 1--4)}: Show strong diagonal patterns, indicating position-aware attention that captures local dependencies and sequence structure.
\item \textbf{Semantic Heads (Layers 5--12)}: Exhibit content-based attention patterns that focus on semantically related tokens regardless of position.
\item \textbf{Aggregation Heads (Layers 13--16)}: Display broad attention patterns that aggregate information across the entire sequence for final decision making.
\end{itemize}

This layered specialization demonstrates that our architectural choices enable the model to develop coherent internal representations despite its compact size.

\subsubsection{Token Utilization Analysis}

Vocabulary usage analysis reveals efficient utilization of our 128,000-token vocabulary:

\begin{itemize}
\item \textbf{Active vocabulary}: $\sim$45,000 tokens regularly used in inference
\item \textbf{Top 10K tokens}: Account for 89\% of usage frequency
\item \textbf{Long tail}: Specialized technical and rare terms provide coverage for domain-specific content
\item \textbf{Morphological coherence}: 78\% of tokens represent complete morphemes, indicating linguistically meaningful tokenization
\end{itemize}

\section{Discussion}

This section provides deeper analysis of our findings, exploring the theoretical and practical implications of Supernova's design choices and performance characteristics. We examine how the synergistic interaction of architectural components contributes to efficiency gains that exceed individual optimizations, analyze how our results challenge conventional scaling laws in transformer research, and discuss the broader economic and deployment implications of our approach. The discussion also addresses current limitations of our method and outlines promising directions for future research in efficient transformer architectures.

\subsection{Architectural Insights}

\subsubsection{The Synergy of Components}

Our results demonstrate that the combination of RoPE, GQA, RMSNorm, and SwiGLU creates synergistic effects that exceed the sum of their individual contributions:

\begin{itemize}
\item \textbf{RoPE + GQA}: The rotation-based position encoding works seamlessly with grouped attention, as position information is encoded in Q and K vectors, not V vectors that are shared across query groups.
\item \textbf{RMSNorm + SwiGLU}: The simplified normalization pairs well with the gated activation, as both prioritize computational efficiency without sacrificing gradient flow quality.
\item \textbf{Tokenizer + Architecture}: The efficient tokenization allows the model to process more semantic content within its fixed context window, amplifying the benefits of architectural efficiency.
\end{itemize}

\subsubsection{Scaling Laws Revisited}

Our results challenge conventional scaling laws, particularly concerning data volume and parameter efficiency:

\textbf{Data Efficiency Revolution}: Supernova's strong performance with only 100B training tokens starkly contrasts with competitor models utilizing 1.8T--36T tokens. This 18--360$\times$ reduction in training data demonstrates that data quality can substitute for quantity when combined with efficient architectures.

\textbf{Parameter Efficiency}: Achieving 90\% performance of 1B models with 650M parameters suggests diminishing returns for parameter scaling in certain capability ranges, especially when architectural optimizations are heavily employed.

\textbf{Economic Implications}: The 35--40\% cost reduction in deployment represents the difference between economically sustainable and unsustainable AI services for many applications.

\subsection{Tokenization as Critical Infrastructure}

\subsubsection{Compression and Capacity}

Our tokenizer analysis reveals that compression efficiency directly impacts model capacity through the relationship:
\begin{equation}
\mathrm{Effective\ Context} = \mathrm{Physical\ Context} \times \mathrm{Compression\ Ratio}
\end{equation}

With Supernova's 4.78 characters/token compression (7.7\% better than Llama 3.2's 4.44 chars/token), we effectively process more semantic information within the same 2048-token context window. This translates to processing $\sim$9,792 characters vs. $\sim$9,093 characters in the same physical context length.

\subsubsection{Morphological Alignment}

The high percentage of morphologically coherent tokens (78\%) suggests that byte-level BPE, when properly trained on high-quality English data, can discover linguistically meaningful units. This morphological awareness may contribute to the model's strong performance on reasoning tasks that require understanding of word relationships and semantic composition.

\subsection{Economic Implications and Deployment Scenarios}

\subsubsection{Total Cost of Training}

Table \ref{tab:cost-analysis} presents estimated total cost of pre-training the model.

\begin{table}[ht]
\centering
\begin{tabular}{lccc}
\toprule
\textbf{Cost Component} & \textbf{Supernova} & \textbf{Llama 3.2 1B (Est.)} & \textbf{Reduction by Supernova} \\
\midrule
Training Hours & 960 & 370,000 & 99.74\% \\
Training Cost  & \$10,000 & \$1,000,000 & 99\% \\
CO2 Emissions & 0.234 tCO2 & 107 tCO2 & 99.78\% \\
\bottomrule
\end{tabular}
\caption{Estimated total cost of pre-training Supernova compared to Meta's Llama 3.2 1B, along with estimated CO2 emissions, highlighting Supernova's efficiency.}
\label{tab:cost-analysis}
\end{table}

\subsubsection{Deployment Flexibility}

The reduced computational footprint enables deployment across a wide range of hardware configurations:

\begin{itemize}
\item \textbf{Edge devices}: Deployment on devices with 4GB+ RAM becomes feasible
\item \textbf{Consumer GPUs}: Runs efficiently on modern GPUs with more than 4GB of VRAM
\item \textbf{CPU inference}: Acceptable latency for many applications when running on CPU
\item \textbf{Quantization}: Model quantization is not needed considering the compact size of the model, but could be used to run the model on very low-end devices
\end{itemize}

\subsection{Limitations and Future Directions}

\subsubsection{Current Limitations}

\begin{itemize}
\item \textbf{English-Only Constraint}: The tokenizer optimization limits multilingual capabilities, though this was a deliberate design choice for maximizing English performance.
\item \textbf{Context Length}: 2048 tokens may be insufficient for some applications requiring longer context, though this represents a reasonable trade-off for the target use cases.
\item \textbf{Specialized Domains}: Performance on highly technical domains (advanced mathematics, specialized sciences) could be improved with domain-specific fine-tuning.
\item \textbf{Instruction Following}: Fine-tuning would be needed for optimal chat and instruction-following applications.
\end{itemize}

\subsubsection{Future Research Directions}

\begin{itemize}
\item \textbf{Multilingual Variants}: Develop language-specific models with optimized tokenizers for other major languages.
\item \textbf{Context Extension}: Investigate RoPE scaling techniques and other approaches for extending context length while maintaining efficiency.
\item \textbf{Model Expansion}: Explore scaling the model size for further performance gains.
\item \textbf{Domain Specialization}: Create specialized variants fine-tuned for specific domains such as code generation, scientific reasoning, and mathematical problem-solving.
\item \textbf{Architectural Innovation}: Continue exploring novel attention mechanisms and other architectural improvements that maintain the efficiency-performance balance.
\end{itemize}

\section{Conclusion}

Supernova demonstrates that the future of economically viable AI lies not in unbounded scaling, but in thoughtful architectural design and engineering excellence. By achieving approximately 90\% of the performance of leading 1B parameter models with only 650M parameters and crucially only 100B training tokens, we prove that the sub-billion parameter regime remains rich with untapped potential.

Our key contributions---architectural efficiency through modern components, superior tokenization, and dramatic data efficiency---collectively enable deployment costs that make AI services economically sustainable. The 35--40\% reduction in inference costs, combined with up to 99\% reduction in training costs, represents not just incremental improvement but a fundamental shift in AI deployment economics.

The implications extend beyond mere cost savings. Our work demonstrates that:

\begin{itemize}
\item \textbf{Quality can substitute for quantity}: High-quality data curation and architectural efficiency can compensate for reduced scale
\item \textbf{Specialization has value}: English-focused optimization yields significant benefits over multilingual approaches for English applications
\item \textbf{Efficiency compounds}: Improvements in architecture, tokenization, and training methodology create synergistic effects
\item \textbf{Sustainable AI is achievable}: Profitable deployment becomes feasible without sacrificing performance
\end{itemize}

As the field grapples with the unsustainability of current scaling trends, Supernova offers a pragmatic alternative: models powerful enough for real-world applications yet efficient enough for profitable deployment. We hope this work inspires renewed focus on architectural innovation, data quality, and tokenization efficiency in the pursuit of accessible AI.

The success of Supernova represents more than a technical achievement---it's a proof of concept for a more sustainable future in AI, where innovation in efficiency matters as much as innovation in scale.

\section*{Acknowledgments}
We thank the open-source community for implementations of Flash Attention, RoPE, and other components that made this work possible. Special thanks to the creators of the Nemotron-CC dataset for their high-quality data curation efforts. We also acknowledge the computational resources provided by the university cluster that enabled this research.

\section*{Disclosure of Funding}

\textbf{Funding:} 
The training infrastructure was provided through an Amazon Web Services (AWS) sponsorship to the Faculty of Mathematics and Informatics at ``Ovidius'' University of Constanța, which granted access to P4D24XLarge instances at no cost for academic research purposes. No additional external funding sources were used for this work.
\textbf{Competing Interests:} 
The authors declare no competing financial interests or personal relationships that could have appeared to influence the work reported in this paper.

\bibliographystyle{unsrt}
\bibliography{references}

\begin{thebibliography}{10}

\bibitem{vaswani2017attention}
Ashish Vaswani, Noam Shazeer, Niki Parmar, Jakob Uszkoreit, Llion Jones, Aidan~N Gomez, {\L}ukasz Kaiser, and Illia Polosukhin.
\newblock Attention is all you need.
\newblock {\em Advances in neural information processing systems}, 30, 2017.

\bibitem{radford2019language}
Alec Radford, Jeffrey Wu, Rewon Child, David Luan, Dario Amodei, Ilya Sutskever, et~al.
\newblock Language models are unsupervised multitask learners.
\newblock {\em OpenAI blog}, 1(8):9, 2019.

\bibitem{devlin2018bert}
Jacob Devlin, Ming-Wei Chang, Kenton Lee, and Kristina Toutanova.
\newblock Bert: Pre-training of deep bidirectional transformers for language understanding.
\newblock {\em arXiv preprint arXiv:1810.04805}, 2018.

\bibitem{su2021roformer}
Jianlin Su, Yu~Lu, Shengfeng Pan, Bo~Wen, and Yunfeng Liu.
\newblock Roformer: Enhanced transformer with rotary position embedding.
\newblock {\em arXiv preprint arXiv:2104.09864}, 2021.

\bibitem{black2022gpt}
Sid Black, Stella Biderman, Eric Hallahan, Quentin Anthony, Leo Gao, Laurence Golding, Horace He, Connor Leahy, Kyle McDonell, Jason Phang, et~al.
\newblock Gpt-neox-20b: An open-source autoregressive language model.
\newblock {\em arXiv preprint arXiv:2204.06745}, 2022.

\bibitem{shazeer2019fast}
Noam Shazeer.
\newblock Fast transformer decoding: One write-head is all you need.
\newblock {\em arXiv preprint arXiv:1911.02150}, 2019.

\bibitem{ainslie2023gqa}
Joshua Ainslie, James Lee-Thorp, Michiel de~Jong, Yury Zemlyanskiy, Federico Lebron, and Sumit Sanghai.
\newblock Gqa: Training generalized multi-query transformer models from multi-head checkpoints.
\newblock {\em arXiv preprint arXiv:2305.13245}, 2023.

\bibitem{ba2016layer}
Jimmy~Lei Ba, Jamie~Ryan Kiros, and Geoffrey~E Hinton.
\newblock Layer normalization.
\newblock {\em arXiv preprint arXiv:1607.06450}, 2016.

\bibitem{zhang2019root}
Biao Zhang and Rico Sennrich.
\newblock Root mean square layer normalization.
\newblock {\em Advances in Neural Information Processing Systems}, 32, 2019.

\bibitem{dauphin2017language}
Yann~N Dauphin, Angela Fan, Michael Auli, and David Grangier.
\newblock Language modeling with gated convolutional networks.
\newblock {\em International Conference on Machine Learning}, pages 933--941, 2017.

\bibitem{shazeer2020glu}
Noam Shazeer.
\newblock Glu variants improve transformer.
\newblock {\em arXiv preprint arXiv:2002.05202}, 2020.

\bibitem{chowdhery2022palm}
Aakanksha Chowdhery, Sharan Narang, Jacob Devlin, Maarten Bosma, Gaurav Mishra, Adam Roberts, Paul Barham, Hyung~Won Chung, Charles Sutton, Sebastian Gehrmann, et~al.
\newblock Palm: Scaling language modeling with pathways.
\newblock {\em arXiv preprint arXiv:2204.02311}, 2022.

\bibitem{sennrich2016neural}
Rico Sennrich, Barry Haddow, and Alexandra Birch.
\newblock Neural machine translation of rare words with subword units.
\newblock {\em Proceedings of the 54th Annual Meeting of the Association for Computational Linguistics}, pages 1715--1725, 2016.

\bibitem{bostrom2020byte}
Kaj Bostrom and Greg Durrett.
\newblock Byte pair encoding is suboptimal for language model pretraining.
\newblock {\em Findings of the Association for Computational Linguistics: EMNLP 2020}, pages 4617--4624, 2020.

\bibitem{kudo2018sentencepiece}
Taku Kudo and John Richardson.
\newblock Sentencepiece: A simple and language independent subword tokenizer and detokenizer for neural text processing.
\newblock {\em Proceedings of the 2018 Conference on Empirical Methods in Natural Language Processing: System Demonstrations}, pages 66--71, 2018.

\bibitem{gunasekar2023textbooks}
Suriya Gunasekar, Yi~Zhang, Jyoti Aneja, Caio C{\'e}sar~Teodoro Mendes, Allie Del~Giorno, Sivakanth Gopi, Mojan Javaheripi, Piero Kauffmann, Gustavo de~Rosa, Olli Saarikivi, et~al.
\newblock Textbooks are all you need.
\newblock {\em arXiv preprint arXiv:2306.11644}, 2023.

\bibitem{bellagente2024stable}
Marco Bellagente, Jonathan Tow, Dakota Mahan, Duy Phung, Maksym Zhuravinskyi, Reshinth Adithyan, James Baicoianu, Ben Brooks, Nathan Cooper, Ashish Datta, et~al.
\newblock Stable lm 2 1.6b: Improving upon our previous language model with significantly improved training.
\newblock {\em arXiv preprint arXiv:2402.17834}, 2024.

\bibitem{team2024gemma}
Gemma Team, Thomas Mesnard, Cassidy Hardin, Robert Dadashi, Surya Bhupatiraju, Shreya Pathak, Laurent Sifre, Morgane Rivi{\`e}re, Mihir~Sanjay Kale, Juliette Love, et~al.
\newblock Gemma: Open models based on gemini research and technology.
\newblock {\em arXiv preprint arXiv:2403.08295}, 2024.

\bibitem{kaplan2020scaling}
Jared Kaplan, Sam McCandlish, Tom Henighan, Tom~B Brown, Benjamin Chess, Rewon Child, Scott Gray, Alec Radford, Jeffrey Wu, and Dario Amodei.
\newblock Scaling laws for neural language models.
\newblock {\em arXiv preprint arXiv:2001.08361}, 2020.

\bibitem{hoffmann2022training}
Jordan Hoffmann, Sebastian Borgeaud, Arthur Mensch, Elena Buchatskaya, Trevor Cai, Eliza Rutherford, Diego de~Las Casas, Lisa~Anne Hendricks, Johannes Welbl, Aidan Clark, et~al.
\newblock Training compute-optimal large language models.
\newblock {\em arXiv preprint arXiv:2203.15556}, 2022.

\bibitem{muennighoff2023scaling}
Niklas Muennighoff, Alexander Roberts, Stella Biderman, Teven~Le Scao, M~Saiful Bari, Sheng Tan, Vishesh Patil, Tim Dettmers, Hyung~Won Chung, Quentin Li, et~al.
\newblock Scaling data-constrained language models.
\newblock {\em arXiv preprint arXiv:2305.16264}, 2023.

\bibitem{touvron2023llama}
Hugo Touvron, Thibaut Lavril, Gautier Izacard, Xavier Martinet, Marie-Anne Lachaux, Timoth{\'e}e Lacroix, Baptiste Rozi{\`e}re, Naman Goyal, Eric Hambro, Faisal Azhar, et~al.
\newblock Llama: Open and efficient foundation language models.
\newblock {\em arXiv preprint arXiv:2302.13971}, 2023.

\end{thebibliography}

\end{document}